%% file: main.tex
\documentclass[twoside]{article}

\usepackage[preprint]{aistats2027}

\usepackage[round]{natbib}

\usepackage{amsmath,amssymb,amsthm}
\usepackage{graphicx}
\usepackage{booktabs}
\usepackage[colorlinks,linktoc=all]{hyperref}       
\usepackage[dvipsnames]{xcolor}
\hypersetup{linkcolor=MidnightBlue,citecolor=MidnightBlue,urlcolor=MidnightBlue}
\usepackage[capitalize,noabbrev]{cleveref}

\usepackage{bm}
\usepackage{nicefrac}
\def\vx{{\bm{x}}}

\def\vk{{\bm{k}}}

\begin{document}


\twocolumn[

\aistatstitle{Understanding Hidden Spatial Preferences in Bayesian Optimization}

\aistatsauthor{Maria Bånkestad \And Jens Sjölund}
\aistatsaddress{Research Institutes of Sweden (RISE) \And Uppsala University} ]

\begin{abstract}
Bayesian optimization uses posterior uncertainty to decide where to
explore, but this uncertainty is not spatially neutral. On bounded
domains, Gaussian-process posterior variance is typically larger near faces, edges, and corners, giving variance-based acquisition functions a systematic boundary bias. This spatial preference is usually inherited from the surrogate rather than chosen as part of the exploration strategy.
We separate these roles by keeping the predictive model fixed and
modifying the spatial preference of the exploration term. We control the preference with spatial weight fields and amortize their construction with neural surrogates. Controlled experiments show that the Gaussian process' default boundary bias changes which parts of the domain are explored and which optima are
reached.
We compare three alternative exploration principles and find, across standard benchmark functions and stochastic robot-pushing tasks,
that they can often lower regret, particularly early in the optimization.
The spatial structure of posterior uncertainty is therefore not a
neutral default: it shapes the exploration strategy and can materially
affect optimization performance.
\looseness-1
\end{abstract}

\section{INTRODUCTION}
\label{sec:intro}

\begin{figure}[t!]
\centering
\includegraphics[width=0.93\columnwidth]{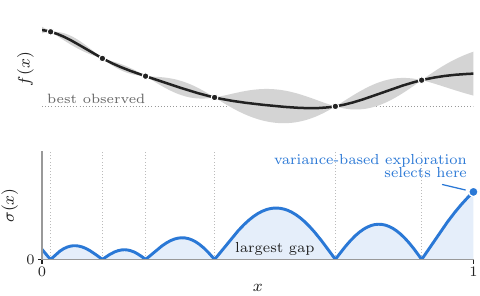}
\caption{A Gaussian process fitted on a bounded domain (top) and its posterior standard deviation (bottom). In this example, uncertainty is highest near the domain boundaries, so pure variance exploration selects a boundary point despite the larger gap between observations in the interior.}
\label{fig:intro}
\end{figure}

In Gaussian-process (GP) Bayesian optimization~\citep{garnett2023bayesian}
(BO), posterior uncertainty influences both how much to explore and where
that exploration is directed. The overall strength of exploration is
controlled by the acquisition function's exploration weight, while its
spatial preference is inherited from the GP posterior. On bounded domains,
posterior variance is typically larger near faces, edges, and corners,
producing a systematic boundary bias in variance-based exploration
(\cref{fig:intro}; \citealp{siivola2018boundary,bankestad2026boundary}).\looseness=-1

This boundary bias is not inherently beneficial or harmful---exploring the boundary helps when the optimum lies there. Standard BO, however, inherits it from the GP surrogate regardless of whether it is appropriate
for the optimization problem. We therefore suggest complementing the familiar question of \emph{how much} to explore with asking \emph{where} to explore.

Previous work has addressed spatial effects in Bayesian optimization, including methods for reducing boundary-seeking behavior
\citep{siivola2018boundary,oh2018bock}. 
In Bayesian experimental design, \citet{jarl2026boundary} proposed a geometric equalizer that removes the
expected spatial variation of GP uncertainty without changing the predictive model. 
Here we extend this idea to BO by treating the spatial preference of variance-based exploration as a design choice. 
We introduce a spatial weight on the exploration term, compare three principles for defining it, and learn reusable weight fields for efficient use during
optimization.\looseness-1

We isolate this effect in a controlled landscape with local minima at different spatial locations, changing only which minimum is global. 
This allows us to test how the spatial preference of exploration affects which regions BO reaches. 
The default rule favors corner minima, whereas alternative spatial preferences more often reach face and interior minima.
Similar differences appear on standard benchmark functions and stochastic robot-pushing tasks, where spatial weighting often reduces regret early in the optimization.\looseness=-1

\section{BACKGROUND AND RELATED WORK}
\label{sec:background}
We consider Bayesian optimization of an unknown function $f\colon \Omega \to \mathbb{R}$ on the unit cube \(\Omega=[0,1]^D\), given access only to noisy observations $y = f(\vx) + \varepsilon$, $\varepsilon \sim \mathcal{N}(0, \sigma_n^2)$.
As usual, we model the unknown function as a Gaussian process, $f \sim \mathcal{GP}(\mu(\vx), k(\vx, \vx'))$, with mean function $\mu$ and covariance kernel $k$.
Given observations $\mathcal{D}=\{(\vx_i, y_i)\}_{i=1}^N$, the posterior covariance between any two points $\vx$ and $\vx'$ is~\citep{rasmussen2006gaussian}
\begin{equation}
   c(\vx, \vx') = k(\vx, \vx') - \vk_\vx^\top (K + \sigma_n^2 I)^{-1} \vk_{\vx'},
\end{equation}
where the matrix elements $K_{ij} = k(\vx_i, \vx_j)$, and $\vk_\vx = (k(\vx_1, \vx), \ldots, k(\vx_N, \vx))^\top$.
In particular, the posterior variance at $\vx$ is the diagonal entry $\sigma^2(\vx) = c(\vx, \vx)$.

Bayesian optimization primarily targets settings where observations are costly to acquire.
To choose the next evaluation, it combines a predictive model (the GP)
with an acquisition function~\(\alpha:\Omega\to\mathbb{R}\) that scores
candidate locations \(\vx\).
Common acquisition functions include expected improvement \citep{jones1998efficient} and confidence-bound methods \citep{srinivas2010gpucb}. 
For minimization, we focus on the lower-confidence-bound acquisition
\begin{equation}
\alpha_N(\mathbf{x})
=
\mu_N(\mathbf{x})
-
\lambda\,\sigma_N(\mathbf{x}),
\end{equation}
where \(\mu_N(\mathbf{x})\) and \(\sigma_N(\mathbf{x})\) denote the GP posterior mean and standard deviation, respectively, and \(\lambda\) controls the strength of exploration relative to exploitation.

\begin{figure}[t]
    \centering
    \includegraphics[width=\columnwidth]{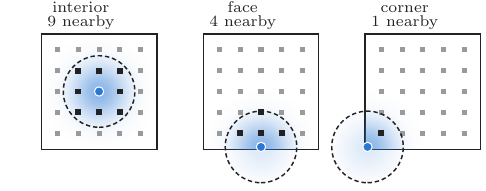}
    \caption{
    Near edges and corners, fewer observations lie within the effective correlation neighborhood, leading to higher posterior uncertainty.
    }
    \label{fig:boundary_geometry}
\end{figure}

Let \(X_N=\{\mathbf{x}_i\}_{i=1}^N\) denote the current design, that is, the \(N\) locations observed so far. For fixed kernel hyperparameters, \(\sigma_N(\mathbf{x})\) depends on \(X_N\), the kernel, and the noise model, but not on the observed function values (see, e.g., \citet{krause2008near}). Consequently, spatial variation in posterior uncertainty induces a spatial preference in variance-based exploration \citep{bankestad2026boundary}. On bounded domains, uncertainty is often higher near the boundary because locations there have fewer neighboring observations within the domain that can reduce it; \cref{fig:boundary_geometry} illustrates this effect.

Several methods have addressed related boundary effects by changing the
predictive model or its geometry. \citet{siivola2018boundary} introduce
virtual derivative-sign observations under an interior-optimum
assumption, while BOCK \citep{oh2018bock} replaces the kernel geometry
with a cylindrical representation that reduces boundary attraction.
Input transformations provide another way to alter the geometry seen by
the surrogate \citep{snoek2014inputwarping,doumont2026}. In these
approaches, changes in exploration behavior are coupled to changes in
the predictive model or its geometry.\looseness-1

Spatial preferences can also be introduced directly through the
acquisition function. \(\pi\)BO \citep{hvarfner2022pibo} reweights the
acquisition using a user-specified prior over the location of the
optimum. In contrast, the preference studied here arises from the
surrogate uncertainty itself, and we modify only the exploration term.

\subsection{Alternative Exploration Principles}
\label{sec:background_where}

Local posterior variance is not the only way to decide where an observation is useful. In optimal experimental design and active learning, a candidate can instead be valued by how much it is expected to reduce uncertainty about the model or its predictions \citep{mackay1992information,cohn1996active}. For Gaussian processes, integrated variance reduction makes this idea explicit by favoring points whose observation reduces posterior uncertainty over a large part of the domain, even when their local variance is not maximal. Information-theoretic BO applies a related principle to information about the optimizer \citep{hennig2012entropy,hernandezlobato2014pes}, while prediction-oriented active learning targets information about predictions in a region of interest \citep{bickfordsmith2023epig}. This suggests an alternative exploration principle: favor locations expected to reduce uncertainty elsewhere.

A different principle is to maximize spatial coverage. Minimax and maximin designs promote coverage and separation \citep{johnson1990minimax}; Latin-hypercube designs are widely used for computer experiments \citep{morris1995exploratory}; and low-discrepancy sequences such as scrambled Sobol points provide approximately uniform coverage of the domain \citep{sobol1967distribution,owen1998scrambling}. These methods target coverage of the accumulated design, whereas our uniform-allocation principle targets the distribution of the next pure-exploration selection.

\section{METHODS}
\label{sec:methods}

\label{sec:background_gap}
Standard GP-based BO uses the same posterior for prediction and exploration, but these roles need not be coupled. 
In Bayesian experimental design, \citet{jarl2026boundary} showed that the GP's boundary bias can be modified while leaving the predictive model unchanged.
We adopt the same separation in Bayesian optimization and treat the spatial preference of exploration as a design choice.

\subsection{Controlling Spatial Preferences by Weighting}
\label{sec:spatial_allocation}

To control the spatial
preference of exploration, we introduce a weight field
\(g:\Omega\to\mathbb{R}_+\) that acts on the uncertainty term of the
acquisition function:
\begin{equation}
\alpha_g(\mathbf{x})
=
\mu_N(\mathbf{x})
-
\lambda\,
g(\mathbf{x})\,
\sigma_N(\mathbf{x}).
\end{equation}
The choice \(g(\mathbf{x})=1\) recovers standard GP-LCB. Changing \(g\) changes where exploration is emphasized.

We consider three principles for choosing the weight field:
equalizing systematic spatial variation in expected uncertainty,
making pure-exploration selections approximately uniform over the
domain, and favoring locations whose uncertainty produces greater
domain-wide uncertainty reduction.

\subsection{Reference Designs and Expected Uncertainty}
\label{sec:reference_designs}

Let \(\widetilde X_N\) denote the first \(N\) points of a randomized
Sobol sequence on \(\Omega\). We denote by \(\sigma(\mathbf{x}\mid \widetilde X_N)\) the corresponding posterior standard deviation under a stationary kernel with lengthscale \(\ell\) and fixed observation-noise variance \(\epsilon\).

Following \citet{jarl2026boundary}, we average over reference designs to
isolate the systematic position-dependent component of uncertainty:
\begin{equation}
m(\mathbf{x};N,\ell,D)
=
\mathbb{E}_{\widetilde X_N}
\left[
\sigma(\mathbf{x}\mid \widetilde X_N)
\right].
\end{equation}
On a bounded domain, \(m(\mathbf{x};N,\ell,D)\) generally varies with
position. The three weight fields use this expected spatial structure in
different ways.

\subsection{Three Exploration Principles}
\label{sec:allocation_principles}
\begin{figure*}[t]
\centering
\includegraphics[width=0.93\textwidth]{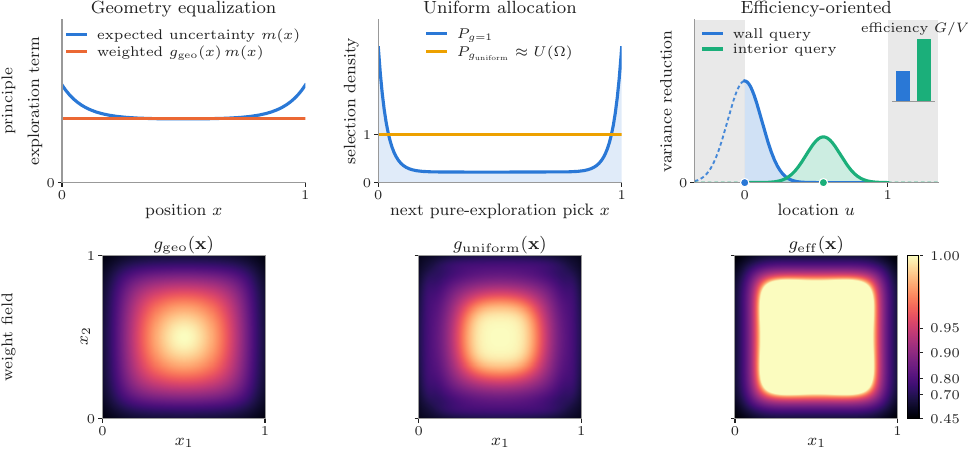}
\caption{The three exploration principles. Top: the quantity targeted by
each principle in one dimension. Geometric equalization flattens the
expected weighted exploration term, uniform allocation targets the
distribution of the next pure-exploration selection, and
efficiency-oriented exploration favors locations whose uncertainty
produces greater reduction across the domain. Bottom: the corresponding
weight fields in \(D=2\) for \(N=10\) and \(\ell_e=0.27\), shown on a
common scale with \(g=1\) denoting no reweighting.}
\label{fig:principles}
\end{figure*}

We now define the three weight fields. \Cref{fig:principles} illustrates what each principle targets and shows the resulting weight fields in two dimensions.
Each field depends on the number of observations \(N\), lengthscale
\(\ell\), and dimension \(D\); we omit this dependence from the notation
where it is clear from context.

\paragraph{Geometric equalization.}

The first principle compensates for the systematic spatial variation of
posterior uncertainty. Let
\(\mathbf{x}_c=(1/2,\ldots,1/2)\) denote the center of the domain. We
define
\begin{equation}
g_{\mathrm{geo}}(\mathbf{x})
=
\frac{
m(\mathbf{x}_c;N,\ell,D)
}{
m(\mathbf{x};N,\ell,D)
}.
\end{equation}
By construction,
\begin{equation}
g_{\mathrm{geo}}(\mathbf{x})
m(\mathbf{x};N,\ell,D)
=
m(\mathbf{x}_c;N,\ell,D).
\end{equation}
Locations with systematically higher uncertainty due to their position are
therefore downweighted, while locations with lower expected uncertainty
are upweighted (\cref{fig:principles}, left). This is the geometric
equalizer introduced by \citet{jarl2026boundary} for Bayesian experimental
design; here we use the same principle to control exploration in Bayesian optimization.

\paragraph{Uniform allocation.}

Geometric equalization makes the weighted expected uncertainty constant
across the domain, but this does not imply that pure exploration selects
all locations equally often. Equalizing the expected exploration value
is different from equalizing the distribution of its maximizer. The second principle instead targets the selected locations directly.

For any weight field \(g\), define the next pure-exploration selection
under a reference design \(\widetilde X_N\) as
\begin{equation}
\mathbf{x}_g(\widetilde X_N)
=
\arg\max_{\mathbf{x}\in\Omega}
g(\mathbf{x})
\sigma(\mathbf{x}\mid \widetilde X_N).
\end{equation}
We denote by \(P_g^{N,\ell,D}\) the distribution of these selections
induced by the reference-design distribution. The uniform-allocation
field \(g_{\mathrm{uniform}}\) is chosen so that this selection
distribution is approximately uniform over \(\Omega\). The target
therefore concerns the next pure-exploration selection across reference
designs, rather than the accumulated design
(\cref{fig:principles}, middle).

\paragraph{Efficiency-oriented exploration.}
The third principle measures the domain-wide uncertainty reduction
associated with an evaluation. For a reference design \(\widetilde X_N\), let \(c_{\widetilde X_N}(\mathbf{u},\mathbf{x})\) denote the posterior covariance between \(\mathbf{u}\in\Omega\) and a candidate location \(\mathbf{x}\in\Omega\). The integrated variance reduction from
\(\mathbf{x}\) is~\citep{seo2000nipv}
\begin{equation}
G_{\widetilde X_N}(\mathbf{x})
=
\frac{1}{|\Omega|}
\int_\Omega
\frac{
c_{\widetilde X_N}(\mathbf{u},\mathbf{x})^2
}{
\sigma^2(\mathbf{x}\mid\widetilde X_N)+\epsilon
}
\,d\mathbf{u}.
\end{equation}
Here we average posterior variance rather than the posterior standard
deviation used in \(m\), since \(G\) measures variance reduction.
Averaging over reference designs gives
\begin{equation}
V(\mathbf{x})
=
\mathbb{E}_{\widetilde X_N}
\left[
\sigma^2(\mathbf{x}\mid\widetilde X_N)
\right],
\quad
G(\mathbf{x})
=
\mathbb{E}_{\widetilde X_N}
\left[
G_{\widetilde X_N}(\mathbf{x})
\right].
\end{equation}
We choose the square root so that the squared weighted uncertainty,
\(g_{\mathrm{eff}}(\mathbf{x})^2 V(\mathbf{x})\), is proportional to
the expected domain-wide variance reduction \(G(\mathbf{x})\). We define
\begin{equation}
g_{\mathrm{eff}}(\mathbf{x})
=
\left(\frac{
G(\mathbf{x})/V(\mathbf{x})
}{
G(\mathbf{x}_c)/V(\mathbf{x}_c)
}\right)^{1/2}.
\end{equation}
Dividing by \(V\) measures domain-wide uncertainty reduction relative to
the local uncertainty available at the candidate. The field therefore
favors locations whose local uncertainty is more effective at reducing
uncertainty across the domain
(\cref{fig:principles}, right).

\subsection{Using the Exploration Principles in Bayesian Optimization}
\label{sec:bo_deployment}
To apply the weight fields in BO, we distinguish the scale used to model
the objective from the scale used to define exploration. Because a global
rescaling of \(g\) changes the overall exploration strength but not its
spatial preference, we normalize the field before applying it:
\begin{equation}
\mathbf{x}_{N+1}
=
\underset{\mathbf{x}\in\Omega}{\arg\min}
\left[
\mu_N(\mathbf{x};\hat{\ell}_N)
-
\lambda\,
\frac{g(\mathbf{x})}{\bar g_N}
\sigma_N(\mathbf{x};\ell_e)
\right].
\end{equation}
Here, \(\hat{\ell}_N\) is the fitted predictive lengthscale and
\(\ell_e\) is the exploration lengthscale. The posterior mean is
evaluated at \(\hat{\ell}_N\), while both \(g\) and the exploration
uncertainty \(\sigma_N\) are evaluated at \(\ell_e\).

In the controlled experiment, \(\bar g_N\) is the domain average,
\begin{equation}
\bar g_N
=
\frac{1}{|\Omega|}
\int_\Omega g(\mathbf{u})\,d\mathbf{u}.
\end{equation}
For the benchmark experiments, we instead average \(g\) over the observed
locations to match the realized exploration strength across rules;
Appendix~\ref{app:benchmark} compares the two normalizations.

In the main comparisons, we hold \(\ell_e\) fixed and vary only \(g\);
all other components of BO, including the predictive GP, hyperparameter
fitting, initial design, exploration weight, and acquisition optimizer,
are shared across methods.

\subsection{Amortized Weight Fields}
\label{sec:numerical_realization}

Recomputing the weight fields as \(N\) and \(\ell\) change during BO would be expensive.
We therefore train principle-specific neural surrogates for each problem
dimension \(D\) over a range of \((N,\ell)\).
Training can be done offline because the fields depend on \(\mathbf{x}\), \(N\), \(\ell\), and \(D\), but not on the observed values or on the realized BO design.
During optimization, the trained surrogate is fixed and is evaluated
at the current \(N\) and chosen exploration lengthscale.
The neural representation also provides gradients with respect to \(\mathbf{x}\) when optimizing the acquisition function.

Each surrogate takes \((\mathbf{x},N,\ell)\) as input. The weight fields
are invariant to reflections about the center of each coordinate axis
and to permutations of the axes, so we encode these symmetries directly
in the architecture. Each coordinate is transformed as
\begin{equation}
    \delta_j = 2x_j(1-x_j),
\end{equation}
which is unchanged under \(x_j\mapsto 1-x_j\). The transformed
coordinates are passed through a shared encoder and summed, making the
output invariant to permutations of the coordinate axes.

The surrogates approximate the quantities used to construct the three weight fields. 
Their training targets, calibration
procedures, architectures, training ranges, Monte Carlo sample counts,
and numerical validation are given in the appendix.

\section{RESULTS}
\label{sec:results}

We first use a controlled synthetic landscape to isolate how spatial
weighting changes where BO searches and which optimum it reaches. We
then test whether the same effects appear on standard benchmark functions
and stochastic robot-pushing tasks.

\subsection{Spatial Preferences Change Where BO Explores}
\label{sec:controlled_bias}

\begin{figure}[t]
\centering
\includegraphics[width=0.5\columnwidth]{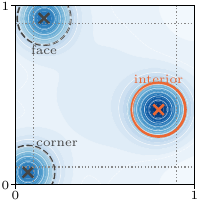}
\caption{The controlled landscape in \(D=2\), with local minima of
identical shape near a corner, a face, and the interior. One minimum is
made deeper and serves as the global minimum. Circles mark the
neighborhoods used to determine whether each minimum is reached, and
dotted lines separate the spatial classes used to categorize evaluation
locations.}
\label{fig:mb4_landscape}
\end{figure}

We first consider the synthetic objective function in Figure~\ref{fig:mb4_landscape}.
It has minima with identical local shape on a weak periodic background:
one near a corner, one near a face, one in the interior, and, for
\(D\ge3\), one near an edge.
(The precise definition is given in Appendix~\ref{app:synthetic}.)
The boundary minima are at the same distance from their nearest wall and
differ in the number of walls they touch. All four have the same objective value except
the designated global minimum, which is lower. 
By changing only which minimum is global, we can test whether the spatial preference favors different parts of the domain.\looseness-1

\begin{figure*}[t!]
\centering
\includegraphics[width=0.95\textwidth]{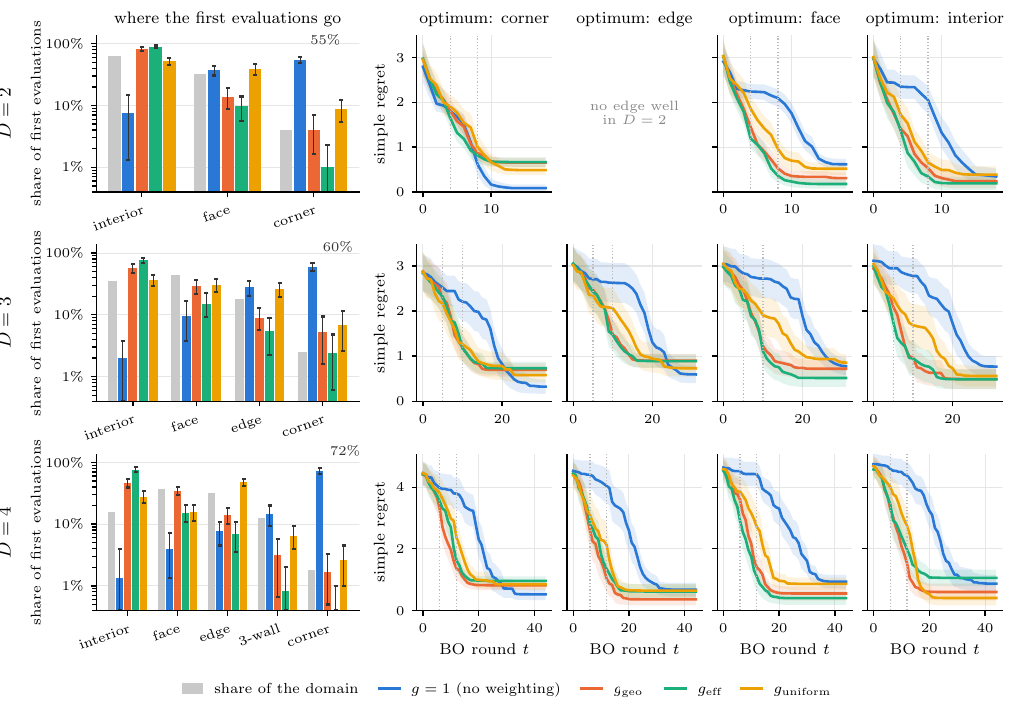}
\caption{Effect of the spatial exploration preference on sampling and
optimization. Each row shows one dimension, with \(\lambda=4\) and
\(25\) paired seeds. Left: fraction of the first \(D+2\) evaluations after
the initial design that fall in the interior, on a face, on an edge, or
near a corner, together with the corresponding fraction of domain volume
in grey. Whiskers are \(90\%\) bootstrap intervals over seeds and the
vertical axis is logarithmic. Right: simple regret per BO round for each
location of the optimum, shown as the mean with \(90\%\) confidence bands.
Vertical dotted lines mark \(t=D+2\) and \(t=2(D+2)\).}
\label{fig:mb4_combined}
\end{figure*}

We compare the default unweighted rule with the three alternative exploration rules at a fixed exploration lengthscale. 
The four methods thus differ only
in terms of their weight fields. In \cref{fig:mb4_combined}, we report results for \(\lambda=4\), for
which the exploration term changes the selected location in about half
of the BO rounds. We use
\(25\) paired seeds for each combination of dimension and global-minimum
location, shared across the four exploration rules. Budgets, results for the remaining values
of \(\lambda\), and standard GP-LCB as a practical reference are given in
Appendix~\ref{app:synthetic}.

We first record where the first \(D+2\) evaluations of each run fall, grouping points by whether they lie in the interior or near one, two, or more walls, and compare these shares with the corresponding share of domain volume (\cref{fig:mb4_combined}, left). 
The unweighted rule strongly prefers corners. In \(D=3\), corners make up only \(2.5\%\) of the domain
but receive \(60\%\) of the first five evaluations, while the interior occupies \(35\%\) of the domain and receives \(2\%\). The corner share increases with dimension, from \(55\%\) in \(D=2\) to \(72\%\) in \(D=4\).
The three spatially weighted rules are much closer to the domain volume, with \(g_{\mathrm{uniform}}\) closest to uniform allocation and
\(g_{\mathrm{eff}}\) shifted somewhat toward the interior. 
This pattern is unchanged by which minimum is global, consistent with
it arising from the exploration term rather than the objective.

These differences in spatial allocation affect how often the rules reach minima in different parts of the domain. For each placement of the global minimum, we record whether a run enters a predefined neighborhood around it within the budget (\cref{tab:mb4_firsthit,tab:mb4_hitfrac} in Appendix~\ref{app:synthetic}). 
When the global minimum is at a corner, the unweighted rule reaches it much more often than the spatially weighted rules (\(83\%\) versus \(17\)--\(33\%\)
in \(D=3\)). When it lies on a face or in the interior, the pattern
reverses; for the interior minimum in \(D=3\), the corresponding rates are
\(19\%\) and \(50\)--\(69\%\). The edge case lies between these two
patterns.

\begin{figure*}[!t]
\centering
\includegraphics[width=\textwidth]{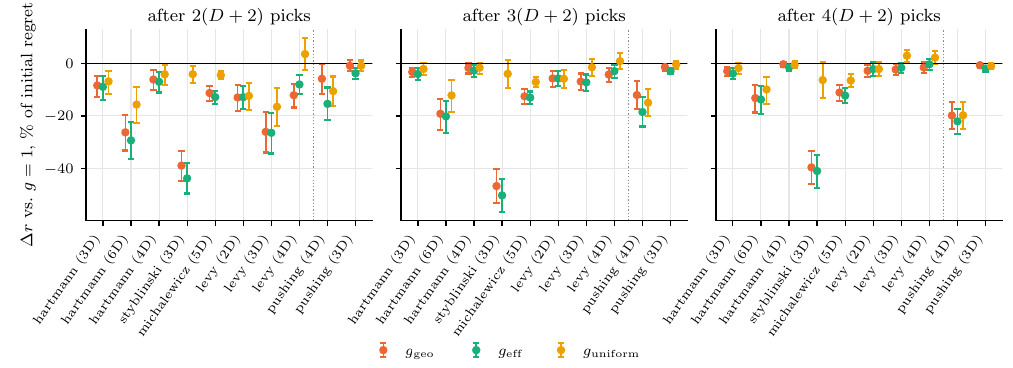}
\caption{Regret difference to the unweighted rule on the benchmark tasks, after \(2(D+2)\), \(3(D+2)\), and \(4(D+2)\) evaluations (left to right), as a percentage of the regret shared by all rules after the initial design; negative values indicate lower regret than \(g=1\). Tasks are ordered by the distance of the optimum to the boundary, with the dimension beside the name; the two robot-pushing simulators are shown to the right of the divider. Points are means over \(100\) paired seeds, with \(90\%\) bootstrap intervals over the paired differences. All rules use \(\lambda=2\) and the same exploration lengthscale. Beale, which every rule solves to zero regret, is omitted.}
\label{fig:bench_summary}
\end{figure*}

\Cref{fig:mb4_combined}, right, shows how these differences in discovery
translate into simple regret. Early in the run, the spatially weighted
rules have lower regret whenever the optimum is not in a corner and, in
\(D=3\) and \(4\), even in the corner case, by about \(1\)--\(3\)
standardized objective units. The corner case also shows that reaching the correct region first
is not the same as resolving the optimum. The unweighted rule enters the corner region early, but its evaluations
concentrate on the corners themselves, where the objective remains above
the nearby minimum. It then
continues to other corners before refining the most promising one.

By the end of the budget, the pattern has changed. The unweighted rule
performs best when the optimum is in a corner, whereas the spatially
weighted rules retain lower regret for face and interior optima and, in
\(D=4\), for the edge optimum as well. The spatial preference of exploration therefore
affects both which optimum regions are found first and how much early
regret is accumulated before they are refined. The differences increase
with \(\lambda\) and are negligible at \(\lambda=1\), where all four rules
are nearly greedy, consistent with the effect originating in the
exploration term.

\subsection{Spatial Preferences Affect Benchmark Performance}
\label{sec:main_results}

We next test whether the spatial effects observed in the controlled
experiment also appear on standard benchmarks. We use nine synthetic
functions in two to six dimensions and two stochastic robot-pushing
simulators in 3D and 4D~\citep{wang2017mes}. Most optima lie in the
interior, while Hartmann 3D, Hartmann 6D, and the 4D simulator have
optima closer to the boundary.\looseness-1

We compare the same four exploration rules at a common exploration
lengthscale, using \(\lambda=2\) and \(100\) paired seeds.
\Cref{fig:bench_summary} reports regret relative to the unweighted rule
after \(2(D+2)\), \(3(D+2)\), and \(4(D+2)\) evaluations, normalized by
the regret after the shared initial design. Full trajectories and
additional baselines are given in
Appendix~\ref{app:benchmark}.

\begin{figure}[t!]
\centering
\includegraphics[width=\columnwidth]{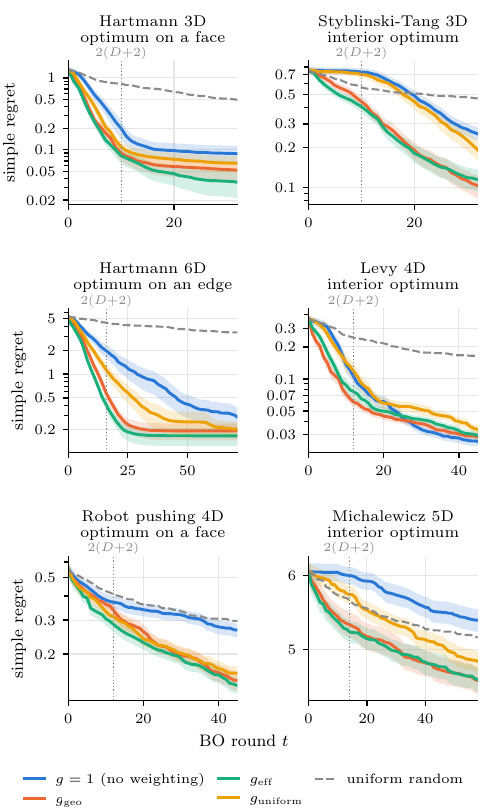}
\caption{Simple regret on six representative tasks. Curves show means
over \(100\) paired seeds with \(90\%\) confidence bands. The dotted
line marks \(2(D+2)\) evaluations, and uniform random sampling is shown
in grey. Full trajectories are given in \cref{app:benchmark}.}
\label{fig:bench_traj}
\end{figure}

After \(2(D+2)\) evaluations, all three weighted rules have lower regret
than the unweighted rule on every task shown in
\cref{fig:bench_summary}. The difference ranges from
less than \(10\%\) of the initial regret on the lower-dimensional
Hartmann functions to about \(40\%\) on Styblinski--Tang. The controlled
experiment suggests one mechanism for this early advantage: the
unweighted rule allocates more evaluations to the boundary, whereas the
weighted rules allocate more to the interior, where most benchmark
optima lie.

By \(3(D+2)\) and \(4(D+2)\) evaluations, the differences have largely
disappeared on Hartmann 3D, Hartmann 4D, and Levy, but remain large on
Styblinski--Tang and visible on Michalewicz and Hartmann 6D
(\cref{fig:bench_summary}). On Hartmann 6D, the weighted rules reach the
region containing the global minimum earlier, while on
Styblinski--Tang the separation persists through much of the run
(\cref{fig:bench_traj}). More generally, performance gaps can persist
when early evaluations lead the rules into different regions, but tend
to shrink once they reach the same region.

The robot-pushing tasks show that the same spatial effects also appear
in stochastic optimization problems. All
four GP rules solve the 3D task within the budget, so their differences
are mainly early. On the 4D task, however, the weighted rules have lower
regret at every checkpoint and finish \(19\)--\(23\%\) below the
unweighted rule relative to the initial regret. The unweighted rule also
allocates many evaluations to the box boundaries. Because the
orientation variable is periodic, two of these boundaries correspond to
the same physical pose. Boundary-seeking exploration can therefore be
costly even when a box boundary is only an artifact of the
parameterization.

Levy shows the opposite late-stage behavior. The unweighted rule spends
more evaluations near the boundary early, increasing regret at first but
also reducing uncertainty there. The weighted rules postpone those
evaluations and retain more boundary uncertainty later. Their early
advantage therefore disappears, and the unweighted rule finishes
marginally ahead.

Michalewicz exposes a different limitation. With the common exploration
lengthscale, all four rules, including \(g=1\), finish close to uniform
random sampling, while standard GP-LCB with the fitted lengthscale
performs substantially better (Appendix~\ref{app:benchmark}). The common
lengthscale is too large to resolve the function's narrow structure, and
spatial reweighting cannot correct this mismatch. Thus \(g\) controls
where exploration is directed, whereas the exploration lengthscale
controls the scale at which uncertainty is resolved.

\section{DISCUSSION}
\label{sec:discussion}

Our main result is that variance-based BO contains an implicit spatial
exploration policy. On bounded domains, this often favors faces, edges,
and corners, and our experiments show that this preference can affect
which regions are reached and which optima are found. The important
point is not that boundary exploration is inherently undesirable, but
that this preference is usually inherited from the GP rather than chosen
for the optimization problem.

There is unlikely to be a universally best spatial preference. Our three principles provide problem-agnostic alternatives when no
reliable prior over the optimum location is available. If such prior
knowledge does exist, it can instead be used to favor particular parts
of the domain. Methods such as \(\pi\)BO
\citep{hvarfner2022pibo} make this type of prior preference explicit by
reweighting the acquisition function. Our setting is complementary:
even without such prior information, the GP already induces a spatial
preference through its uncertainty. Combining explicit prior beliefs
with control of this surrogate-induced preference is a natural direction
for future work.

When little is known about the location of the optimum, uniform
allocation provides a natural reference, but our results show that
spatial neutrality and search efficiency are not the same objective.
Geometric equalization and efficiency-oriented exploration can perform
better while retaining structure in the GP uncertainty rather than
forcing the next exploration location to be uniformly distributed.
The Levy experiments also show that the differences between rules are not determined by iteration number alone. It depends on the remaining spatial structure of uncertainty: evaluations
avoided near the boundary early can leave uncertainty there to be
resolved later.

Spatial weighting addresses only one aspect of exploration. On
Michalewicz, changing \(g\) cannot compensate for an exploration
lengthscale that is too large to resolve the narrow structure of the
objective. This separates two choices: \(g\) controls where exploration
is directed, while the exploration lengthscale controls the scale at
which uncertainty is resolved.
The present constructions also exploit the symmetry of a box domain;
irregular, constrained, or periodic domains may require weight fields
adapted to their geometry.

We have focused on GP-LCB because its additive exploration term makes
this separation explicit. Extending the same perspective to other
acquisition functions is less direct, but would help clarify how
surrogate-induced spatial preferences appear in methods such as Thompson
sampling \citep{thompson1933} and Log Noisy Expected Improvement
\citep{ament2023logei}.

\section{CONCLUSION}
\label{sec:conclusion}

Variance-based BO contains an implicit spatial exploration policy. By
separating this policy from the predictive model, we show that changing
where uncertainty is emphasized can alter where BO searches, which optima it finds, and its regret. 
The GP's default preference can be useful near the boundary, while alternative preferences can improve search elsewhere, particularly early in optimization. 
The spatial preference of exploration should therefore be treated as an explicit design choice in BO rather than as an unexamined consequence of the surrogate model.\looseness-1

\clearpage
\section*{ACKNOWLEDGEMENTS}
This work was partially supported by the Wallenberg AI, Autonomous Systems and Software Program (WASP) funded by the Knut and Alice Wallenberg Foundation. The computations were enabled by resources in project UPPMAX 2026/1-226 provided by Uppsala University at UPPMAX.

\bibliography{references}

\clearpage
\appendix
\thispagestyle{empty}

\onecolumn
\aistatstitle{Understanding Hidden Spatial Preferences in Bayesian Optimization: \\
Supplementary Materials}

\section{WEIGHT-FIELD SURROGATES}
\label{app:fields}

This appendix describes how the three weight fields of
\cref{sec:allocation_principles} are constructed and how their neural
surrogates are trained and validated.

\paragraph{Reference designs.}
All expectations are taken over reference designs \(\widetilde X_N\),
the first \(N\) points of a scrambled Sobol sequence, translated by a
uniformly random shift modulo one. The posterior under a reference
design uses a Mat\'ern-\(5/2\) kernel with unit output scale,
lengthscale \(\ell\), and noise variance \(\epsilon=10^{-4}\).

\paragraph{Architecture.}
For each coordinate, the surrogate receives the reflection-invariant
feature \(\delta_j=2x_j(1-x_j)\) together with
\(\log(1+\delta_j/\ell)\). A shared encoder with two hidden layers of
width \(32\) maps these two inputs to a \(32\)-dimensional embedding,
and the embeddings are summed over coordinates. The sum is concatenated
with eight further inputs: \(\log N\) and \(\log\ell\), both
standardized over the training range, \(N^{1/D}\), \(D\), the product
\(P=\prod_j 2\delta_j\), \(\log(1+P/\ell)\), the mean
\(\tfrac{1}{D}\sum_j e^{-\delta_j/\ell}\), and
\(\log\big(1+\sum_j e^{-\delta_j/\ell}\big)\). A network with hidden
widths \(64\), \(64\), and \(32\) maps this vector to a log-score
\(s(\mathbf{x};N,\ell)\). All activations are SiLU. Each field is defined
relative to the center, for example
\(g_{\mathrm{geo}}(\mathbf{x})
=\exp\!\big(s(\mathbf{x}_c;N,\ell)-s(\mathbf{x};N,\ell)\big)\) when
\(s\) represents \(\log m\), so that \(g(\mathbf{x}_c)=1\) holds exactly
and the field is positive and differentiable. One network has about
\(12{,}000\) parameters. We train separate networks for each principle
and each dimension \(D=2,\ldots,6\), over \(N\in[2D,150]\) and
\(\ell\in[0.05,3]\) (\(\ell\in[0.02,3]\) in \(D=2\)). Inputs outside
this range are clamped to it.

\paragraph{Geometric equalization.}
The network represents the expected posterior standard deviation
\(m(\mathbf{x};N,\ell,D)\). It is trained on fresh reference designs at
every step, so no finite training set is reused. Each step draws one
\(N\), \(48\) lengthscales, and query locations concentrated near the
boundary, and uses the average of
\(\sigma(\mathbf{x}\mid\widetilde X_N)\) over \(8\) reference designs as
the regression target. Training runs for \(600{,}000\) steps with a
cosine learning-rate schedule. Both surrogates are trained with Adam, with
the learning rate decayed from \(10^{-3}\) to \(10^{-6}\), and the weights
are averaged over the final \(40\%\) of the steps.

\paragraph{Efficiency-oriented exploration.}
Two networks of the same form represent \(V(\mathbf{x})\) and
\(G(\mathbf{x})\), and \(g_{\mathrm{eff}}\) is formed from their ratio
as in \cref{sec:allocation_principles}. The domain integral in
\(G_{\widetilde X_N}\) is evaluated on \(1024\) quasi-random points,
complemented by \(32\) points around each query location when the
lengthscale is short. Targets are averaged over \(32\) reference designs
for \(G\) and \(64\) for \(V\), and training runs for \(300{,}000\)
steps.

\paragraph{Uniform allocation.}
The uniform-allocation field has no pointwise target, because it is
defined through the distribution of the pure-exploration selection
\(\mathbf{x}_g(\widetilde X_N)\). We calibrate it in two stages.

In the first stage, we fix \((N,\ell)\) at each point of a \(5\times5\)
logarithmic grid over the training range and represent \(\log g\) by the
network above, initialized at \(g_{\mathrm{geo}}\). Each iteration
computes the selections \(\mathbf{x}_g(\widetilde X_N)\) for \(2000\)
reference designs, which are the same in every iteration, using the
acquisition optimizer of \cref{app:optimizer}. A selection is summarized
by its two smallest distances to the walls, each transformed so that it
is uniformly distributed on \([0,1]\) when the selection is uniform on
\(\Omega\). In these coordinates \(\mathbf{v}\), we estimate the ratio
\(\hat r(\mathbf{v})\) between the density of the selections and the
uniform density, using a Gaussian kernel density estimate with bandwidth
\(0.12\) that is reflected at the boundaries. The field is updated as
\begin{equation}
\log g(\mathbf{x})
\;\leftarrow\;
\log g(\mathbf{x})
-\eta\,
\operatorname{clip}\!\big(\log \hat r(\mathbf{v}(\mathbf{x})),\,-0.5,\,0.5\big),
\end{equation}
and the network is refitted to the updated values. The step size starts
at \(\eta=0.1\) and is halved up to four times until the step reduces
the largest Kolmogorov--Smirnov distance between the transformed
coordinates and the uniform distribution. Calibration stops when this
distance, measured on \(1000\) further designs, is within its sampling
variability under uniform selections, when no step has been accepted for
four iterations, or after \(40\) iterations.

In the second stage, a single network is fitted to the \(25\) calibrated
fields by least squares on \(\log g\), giving a field that is continuous
in \((N,\ell)\).

\paragraph{Validation.}
We compare \(g_{\mathrm{geo}}\) and \(g_{\mathrm{eff}}\) against
high-precision Monte Carlo estimates of their definitions that were not
used for training. In the range visited by our experiments, the mean
absolute error is \(0.2\)--\(0.6\%\) for \(g_{\mathrm{geo}}\) and
\(0.6\)--\(0.9\%\) for \(g_{\mathrm{eff}}\). For both fields we train
two networks from independent seeds and use their geometric mean. For
\(g_{\mathrm{uniform}}\), the target can only be met approximately.
When the lengthscale is at least comparable to the design spacing, the
Kolmogorov--Smirnov distance between the distribution of the boundary
distance of the selections and its uniform reference is typically
\(0.1\)--\(0.2\) on held-out \((N,\ell)\) values, compared with
\(0.5\)--\(0.8\) for \(g_{\mathrm{geo}}\) and about \(1\) for the
unweighted rule in \(D\ge3\). For lengthscales well below the design
spacing, no smooth field makes the selections uniform. The invariance to
reflections and permutations holds exactly by construction for all three
fields.

\section{COMMON EXPERIMENTAL SETUP}
\label{app:optimizer}

The following settings are shared by all experiments and all exploration
rules.

\paragraph{Predictive model.}
Objectives are standardized to zero mean and unit variance using
\(8192\) Sobol points. The predictive GP has zero prior mean, a
Mat\'ern-\(5/2\) kernel with unit output scale, and a single
lengthscale \(\hat\ell_N\), refitted in every round by maximum a
posteriori (MAP) estimation under the dimension-scaled log-normal prior
of \citet{hvarfner2024vanilla}. For noise-free objectives, the GP uses a
noise variance of \(10^{-4}\).

\paragraph{Exploration term.}
The unweighted rule and the three weighted rules evaluate both
\(\sigma_N\) and \(g\) at the exploration lengthscale
\(\ell_e=0.268\sqrt{D/2}\), which is fixed throughout a run. We also
report standard GP-LCB, which evaluates \(\sigma_N\) at the fitted
lengthscale \(\hat\ell_N\) without weighting. It differs from the other
rules in the exploration scale as well as in the spatial weighting, and
is therefore not part of the controlled comparison.

\paragraph{Acquisition optimization.}
In every round, the acquisition function is evaluated on \(512\)
scrambled-Sobol candidates. The \(32\) best candidates and \(10\)
further candidates drawn with probability increasing in their
acquisition value serve as starting points for L-BFGS-B, with gradients
propagated through \(g\). The best point found is evaluated.

\paragraph{Initial design, budget, and regret.}
Each run starts from \(2D+2\) scrambled-Sobol points that depend on the
seed and are shared by all rules. The total number of evaluations is
\(B=25\), \(40\), \(55\), \(70\), and \(85\) for \(D=2,\ldots,6\). We
report simple regret in standardized units, with the reference optimum
obtained from \(2^{17}\) Sobol points followed by local optimization.
Comparisons between rules are paired by seed.

\paragraph{Computing infrastructure.}
The weight-field surrogates were trained on single GPUs. The BO runs were
executed on single GPUs or on a CPU cluster with four cores per run.

\section{CONTROLLED SYNTHETIC EXPERIMENT}
\label{app:synthetic}

\paragraph{Objective.}
On the unit cube, the objective is
\begin{equation}
f(\mathbf{x})
=
0.06\,\frac{1}{D}\sum_{j=1}^{D}\cos(2\pi x_j)
-
\sum_{w} a_w
\exp\!\left(-\frac{\|\mathbf{x}-\mathbf{c}_w\|^2}{2s^2}\right),
\end{equation}
with \(a_w=1\) for all Gaussian wells except the one designated to
contain the global minimum, for which \(a_w=1.25\). The well centers
\(\mathbf{c}_w\) and the width \(s\) are as follows.
\(D=2\): corner \((0.07,0.07)\), face \((0.16,0.93)\), interior
\((0.80,0.42)\), \(s=0.100\).
\(D=3\): corner \((0.102,0.898,0.898)\), edge \((0.634,0.898,0.102)\),
face \((0.102,0.384,0.234)\), interior \((0.692,0.292,0.692)\),
\(s=0.146\).
\(D=4\): corner \((0.129,0.129,0.871,0.871)\), edge
\((0.544,0.129,0.344,0.871)\), face \((0.871,0.694,0.294,0.694)\),
interior \((0.367,0.517,0.617,0.367)\), \(s=0.184\).
Let \(h=B^{-1/D}\) denote the spacing of the final design. 
Every boundary well is centered \(0.35\,h\) from its nearest wall, and
the interior well is centered \(1.0\,h\) from every wall. The well with
\(a_w=1.25\) contains the global minimum in every variant, by a margin of
\(0.16\)--\(0.31\) standardized objective units.

\paragraph{Protocol.}
We use \(D=2,3,4\) with the setup of \cref{app:optimizer}, which gives
\(19\), \(32\), and \(45\) BO rounds after the initial design. Each
combination of dimension and global-minimum location is run with
\(25\) seeds and \(\lambda\in\{1,2,4,8\}\). The normalizing average
\(\bar g_N\) is taken over the domain. The exploration lengthscale
corresponds to \(\ell_e/h\) between \(0.67\) at the initial design and
\(1.0\)--\(1.3\) at the end of the run.

\paragraph{Location classes and reached minima.}
We classify a point by the number of its coordinates that lie within
\(h/2\) of a wall: none for the interior, one for a face, two for an
edge, and all \(D\) for a corner. These classes occupy
\(64\%/32\%/4\%\) of the domain in \(D=2\),
\(35\%/44\%/18\%/2.5\%\) in \(D=3\), and
\(16\%/37\%/33\%/13\%/1.8\%\) in \(D=4\). A minimum counts as reached
when an evaluation falls within \(1.5\,s\) of its center, a radius that
contains the corner point itself in every dimension. First-entry
statistics exclude runs in which the initial design already lies in
this neighborhood.

\begin{table}[!htb]
\begin{minipage}[t]{0.48\textwidth}
\centering
\caption{Median number of BO rounds until the neighborhood of the
global minimum is first reached, for \(\lambda=4\); lower is better. An
entry \(>R\) means that the median run did not reach it within the
\(R\) rounds of the budget.}
\label{tab:mb4_firsthit}
\small
\input{figs/tab_mb4_firsthit.tex}
\end{minipage}\hfill
\begin{minipage}[t]{0.48\textwidth}
\centering
\caption{Fraction of runs that reach the neighborhood of the global
minimum within the budget, for \(\lambda=4\).}
\label{tab:mb4_hitfrac}
\small
\input{figs/tab_mb4_hitfrac.tex}
\end{minipage}
\end{table}

\paragraph{Role of the objective.}
The early preference of the unweighted rule for corners does not depend
on which minimum is global. Relative to their share of the domain,
corners are sampled \(13.5\), \(14.5\), and \(13.0\) times more often in
\(D=2\) when the corner, face, or interior minimum is global, and
\(23.0\)--\(25.8\) times more often in \(D=3\). At \(\lambda=4\), the
selected point differs from the minimizer of the posterior mean in
\(37\)--\(63\%\) of the rounds, depending on the rule.

\begin{figure}[!htb]
\centering
\includegraphics[width=0.42\textwidth]{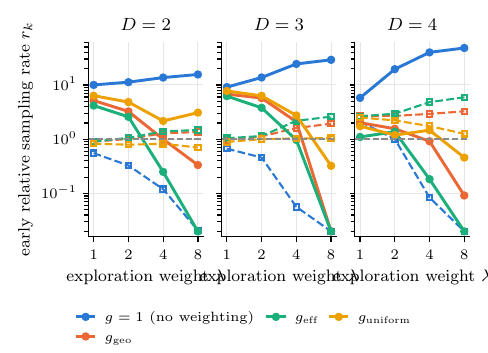}
\caption{Early sampling as a function of the exploration weight
\(\lambda\). The relative sampling rate is the share of the first
\(D+2\) evaluations in a location class divided by the share of the
domain that the class occupies; a value of \(1\) corresponds to uniform
sampling. Solid lines show corners and dashed lines the interior.}
\label{fig:mb4_lambda}
\end{figure}

\paragraph{Dependence on the exploration weight.}
\Cref{fig:mb4_lambda} shows how the early sampling of corners and of the
interior changes with \(\lambda\). At \(\lambda=1\), \(86\)--\(93\%\) of
the selections coincide with the minimizer of the posterior mean, and
the four rules behave similarly. As \(\lambda\) increases, the
unweighted rule moves further into the corners, while the weighted rules
move toward the interior. At \(\lambda=8\), the pattern of the main text
appears with larger margins: the weighted rules reach face and interior
minima within \(3\)--\(8\) rounds, whereas the unweighted rule needs
\(12\)--\(28\) rounds or does not reach them within the budget, and it
remains faster for the corner minimum.

\paragraph{Standard GP-LCB.}
Standard GP-LCB shows the same qualitative boundary preference as the
unweighted rule, with a weaker early corner preference because its
fitted lengthscale is shorter. At \(\lambda=4\), it samples corners
\(12\) and \(8\) times more often than their share of the domain in
\(D=2\) and \(D=3\), compared with \(14\) and \(24\) times for the
unweighted rule.

\section{BENCHMARK EXPERIMENTS}
\label{app:benchmark}

\begin{figure*}[!htb]
\centering
\includegraphics[width=\textwidth]{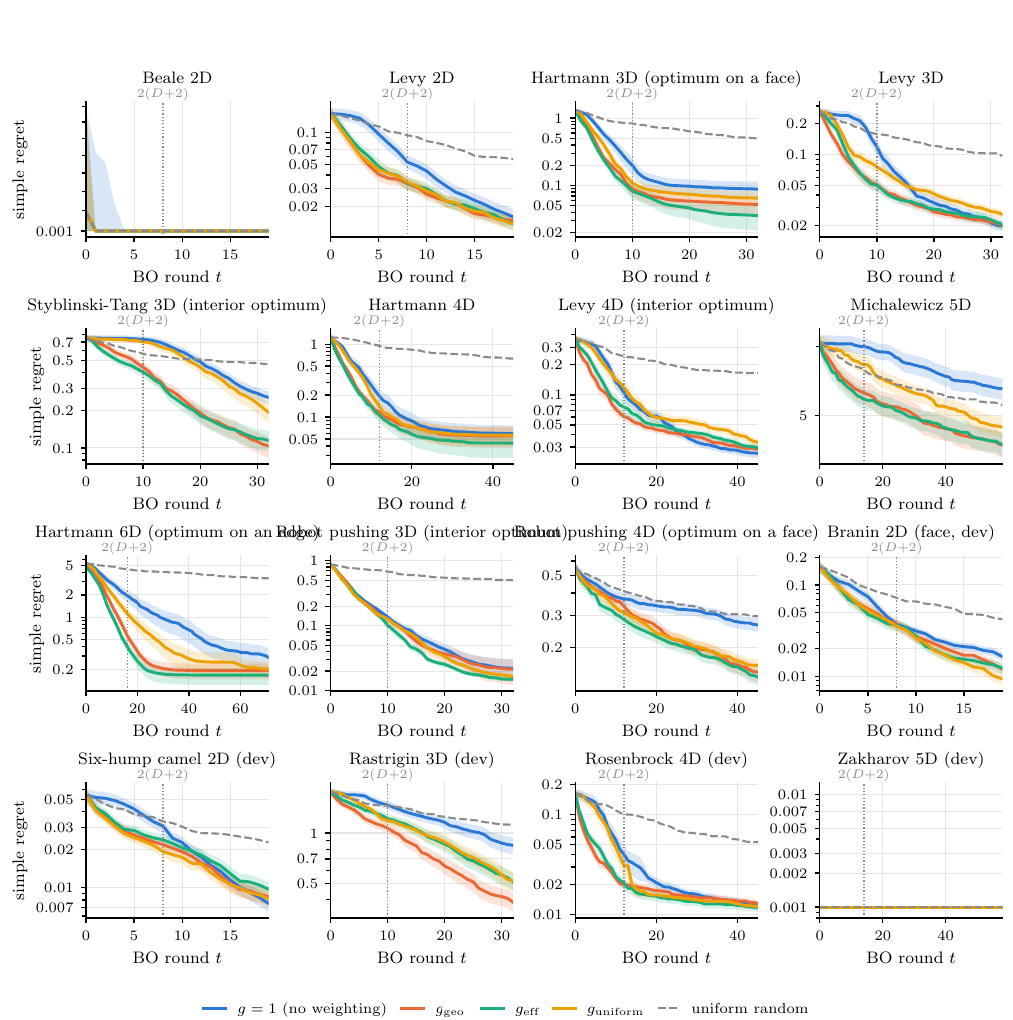}
\caption{Simple regret on every benchmark function, shown as the mean
over \(100\) paired seeds with \(90\%\) confidence bands.}
\label{fig:bench_traj_all}
\end{figure*}

\paragraph{Tasks and protocol.}
The benchmark functions are Beale, Levy 2D, Levy 3D, Hartmann 3D,
Styblinski--Tang 3D, Hartmann 4D, Levy 4D, Michalewicz 5D, and
Hartmann 6D, on their standard domains mapped to \([0,1]^D\). This set
was fixed before any run. Branin, six-hump camel, Rastrigin 3D,
Rosenbrock 4D, and Zakharov 5D were used while developing the method and
are reported separately in \cref{tab:bench_final}. All runs use the
setup of \cref{app:optimizer} with \(\lambda=2\) and \(100\) paired
seeds. For each task, we recorded the location of the global optimum and
its distance to the boundary before the runs. Differences to the
unweighted rule are computed per seed, with \(90\%\)
percentile-bootstrap intervals.
\Cref{fig:bench_traj_all} shows the regret trajectories on every
function.

\begin{table*}[!htb]
\centering\small
\caption{Simple regret at the end of the budget, as the mean and
\(90\%\) bootstrap half-width over \(100\) paired seeds. The columns
\(d_\partial\) and class give the distance of the global optimum to the
boundary and its location class. The column std.\ UCB is standard
GP-LCB, which evaluates the exploration term at the fitted
lengthscale.}
\label{tab:bench_final}
\resizebox{\textwidth}{!}{\input{figs/tab_bench_final.tex}}
\end{table*}

\begin{figure*}[!htb]
\centering
\includegraphics[width=0.9\textwidth]{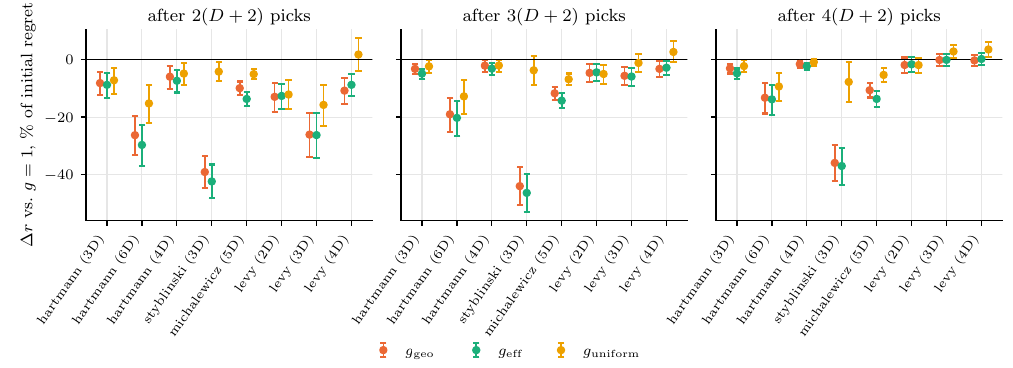}
\caption{As \cref{fig:bench_summary}, with the normalizing average
\(\bar g_N\) taken over the domain instead of over the evaluated
points.}
\label{fig:bench_summary_poolmean}
\end{figure*}

\paragraph{Normalization of the weight field.}
In the benchmark experiments, the normalizing average \(\bar g_N\) is
taken over the evaluated points rather than over the domain. The reason
is exploration strength. With the domain average, the weighted rules
select points at which \(g/\bar g_N\) is \(1.14\)--\(1.19\), because the
boundary is downweighted while the average is fixed at one. At the same
\(\lambda\), they therefore apply a \(15\)--\(20\%\) larger exploration
term at their own selections than the unweighted rule, so that a
comparison at equal \(\lambda\) mixes the spatial preference with the
amount of exploration. Averaging over the evaluated points removes this
difference. We identified the effect in a first set of runs with the
domain average, which \cref{fig:bench_summary_poolmean} reports for
comparison. The early checkpoints are unchanged. At the later
checkpoints, the domain average costs slightly on Levy and gains
slightly on Hartmann 3D and 4D.

\paragraph{Late-stage behavior on Levy.}
On Levy 4D, the unweighted rule places \(67\%\) of its first \(12\)
selections within half a design spacing of a wall, compared with
\(27\%\) for the weighted rules. In the last \(21\) rounds, its share
has fallen to \(32\%\), while that of the weighted rules has risen to
\(41\)--\(49\%\). The weighted rules thus postpone boundary evaluations
rather than avoid them, which is why their early advantage disappears.

\begin{figure}[!htb]
\centering
\includegraphics[width=0.42\textwidth]{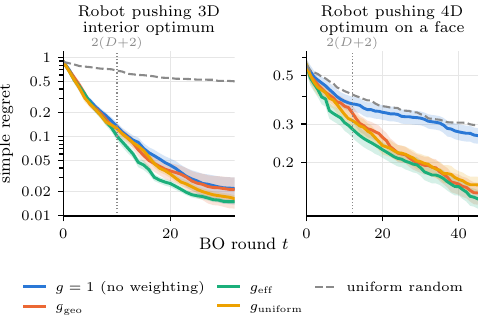}
\caption{Simple regret on the two robot-pushing simulators, shown as the
mean over \(100\) paired seeds with \(90\%\) confidence bands. Regret
below \(0.001\) is drawn at \(0.001\).}
\label{fig:push_traj}
\end{figure}

\paragraph{Robot-pushing simulators.}
The tasks are the 3D and 4D robot-pushing problems of
\citet{wang2017mes}. A pusher starts at a position in
\([-5,5]^2\) and pushes an object for a duration in \([1,30]\); the 4D
task adds the orientation of the pusher in \([0,2\pi]\). The objective
is the distance between the object and a goal after the push. We fix the
goal at \((-4.79,2.67)\), so that every rule and seed optimizes the same
objective. The simulator is stochastic. Each evaluation returns one
simulator run, and regret is measured on the expected objective, which
we estimate from \(100\) runs. The GP noise variance is fixed per task
from repeated runs and equals \(0.3\%\) and \(3.6\%\) of the
standardized variance in 3D and 4D. The optimum lies in the interior in
3D and near a face in 4D. We keep the published box parameterization and
treat the orientation like any other coordinate, although it is
periodic. All other settings are those of the benchmark functions.
\Cref{fig:push_traj} shows the regret trajectories.

\end{document}

%% file: figs/tab_mb4_firsthit.tex
\begin{tabular}{llrrrr}
\toprule
$D$ & optimum at & $g=1$ & $g_{\mathrm{geo}}$ & $g_{\mathrm{eff}}$ & $g_{\mathrm{uniform}}$ \\
\midrule
2 & corner & 2 & $>19$ & $>19$ & $>19$ \\
2 & face & $>19$ & 5 & 4 & 5 \\
2 & interior & 10 & 4 & 4 & 6 \\
\addlinespace
3 & corner & 6 & $>32$ & $>32$ & $>32$ \\
3 & edge & 20 & $>32$ & $>32$ & $>32$ \\
3 & face & $>32$ & 22 & 23 & $>32$ \\
3 & interior & $>32$ & 14 & 5 & 28 \\
\addlinespace
4 & corner & 7 & $>45$ & $>45$ & $>45$ \\
4 & edge & $>45$ & 10 & 13 & $>45$ \\
4 & face & $>45$ & 15 & 8 & $>45$ \\
4 & interior & $>45$ & 14 & $>45$ & 14 \\
\bottomrule
\end{tabular}

%% file: figs/tab_mb4_hitfrac.tex
\begin{tabular}{llrrrr}
\toprule
$D$ & optimum at & $g=1$ & $g_{\mathrm{geo}}$ & $g_{\mathrm{eff}}$ & $g_{\mathrm{uniform}}$ \\
\midrule
2 & corner & 1.00 & 0.06 & 0.28 & 0.33 \\
2 & face & 0.24 & 0.71 & 0.86 & 0.81 \\
2 & interior & 0.68 & 0.89 & 0.95 & 0.74 \\
\addlinespace
3 & corner & 0.83 & 0.17 & 0.17 & 0.33 \\
3 & edge & 0.53 & 0.12 & 0.12 & 0.35 \\
3 & face & 0.22 & 0.50 & 0.50 & 0.28 \\
3 & interior & 0.19 & 0.56 & 0.69 & 0.50 \\
\addlinespace
4 & corner & 0.81 & 0.14 & 0.10 & 0.14 \\
4 & edge & 0.35 & 0.70 & 0.55 & 0.35 \\
4 & face & 0.19 & 0.52 & 0.67 & 0.19 \\
4 & interior & 0.42 & 0.58 & 0.42 & 0.74 \\
\bottomrule
\end{tabular}

%% file: figs/tab_bench_final.tex
\begin{tabular}{lrrlrrrrrr}
\toprule
task & $D$ & $d_\partial$ & class & $g=1$ & std.\ UCB & $g_{\mathrm{geo}}$ & $g_{\mathrm{eff}}$ & $g_{\mathrm{uniform}}$ & random \\
\midrule
\multicolumn{10}{l}{\emph{benchmark suite}}\\
Hartmann 3D & 3 & 0.11 & face & 0.088 $\pm$ 0.025 & 0.044 $\pm$ 0.026 & 0.052 $\pm$ 0.016 & 0.035 $\pm$ 0.014 & 0.065 $\pm$ 0.017 & 0.495 $\pm$ 0.052 \\
Hartmann 6D & 6 & 0.15 & edge & 0.286 $\pm$ 0.085 & 0.153 $\pm$ 0.031 & 0.191 $\pm$ 0.045 & 0.166 $\pm$ 0.043 & 0.201 $\pm$ 0.047 & 3.359 $\pm$ 0.175 \\
Beale 2D & 2 & 0.17 & interior & 0.000 $\pm$ 0.000 & 0.000 $\pm$ 0.000 & 0.000 $\pm$ 0.000 & 0.000 $\pm$ 0.000 & 0.000 $\pm$ 0.000 & 0.000 $\pm$ 0.000 \\
Hartmann 4D & 4 & 0.19 & interior & 0.059 $\pm$ 0.018 & 0.053 $\pm$ 0.017 & 0.055 $\pm$ 0.017 & 0.044 $\pm$ 0.016 & 0.056 $\pm$ 0.017 & 0.635 $\pm$ 0.044 \\
Styblinski-Tang 3D & 3 & 0.21 & interior & 0.253 $\pm$ 0.028 & 0.179 $\pm$ 0.025 & 0.104 $\pm$ 0.020 & 0.115 $\pm$ 0.021 & 0.192 $\pm$ 0.033 & 0.465 $\pm$ 0.029 \\
Michalewicz 5D & 5 & 0.30 & interior & 5.370 $\pm$ 0.150 & 3.224 $\pm$ 0.163 & 4.628 $\pm$ 0.151 & 4.639 $\pm$ 0.138 & 4.854 $\pm$ 0.123 & 5.145 $\pm$ 0.101 \\
Levy 2D & 2 & 0.45 & interior & 0.016 $\pm$ 0.003 & 0.016 $\pm$ 0.002 & 0.015 $\pm$ 0.002 & 0.014 $\pm$ 0.002 & 0.014 $\pm$ 0.002 & 0.056 $\pm$ 0.007 \\
Levy 3D & 3 & 0.45 & interior & 0.020 $\pm$ 0.002 & 0.034 $\pm$ 0.005 & 0.020 $\pm$ 0.002 & 0.020 $\pm$ 0.002 & 0.026 $\pm$ 0.003 & 0.097 $\pm$ 0.009 \\
Levy 4D & 4 & 0.45 & interior & 0.026 $\pm$ 0.003 & 0.037 $\pm$ 0.003 & 0.029 $\pm$ 0.002 & 0.030 $\pm$ 0.003 & 0.034 $\pm$ 0.003 & 0.165 $\pm$ 0.014 \\
\midrule
\multicolumn{10}{l}{\emph{simulators}}\\
Robot pushing 4D & 4 & 0.17 & face & 0.266 $\pm$ 0.022 & 0.243 $\pm$ 0.022 & 0.146 $\pm$ 0.017 & 0.137 $\pm$ 0.013 & 0.159 $\pm$ 0.014 & 0.297 $\pm$ 0.022 \\
Robot pushing 3D & 3 & 0.38 & interior & 0.022 $\pm$ 0.008 & 0.018 $\pm$ 0.001 & 0.021 $\pm$ 0.009 & 0.015 $\pm$ 0.001 & 0.016 $\pm$ 0.002 & 0.500 $\pm$ 0.051 \\
\midrule
\multicolumn{10}{l}{\emph{used during development}}\\
Branin 2D & 2 & 0.04 & face & 0.016 $\pm$ 0.002 & 0.009 $\pm$ 0.002 & 0.012 $\pm$ 0.002 & 0.012 $\pm$ 0.002 & 0.009 $\pm$ 0.001 & 0.042 $\pm$ 0.006 \\
Rosenbrock 4D & 4 & 0.26 & interior & 0.012 $\pm$ 0.001 & 0.010 $\pm$ 0.001 & 0.013 $\pm$ 0.001 & 0.012 $\pm$ 0.001 & 0.012 $\pm$ 0.001 & 0.053 $\pm$ 0.005 \\
Six-hump camel 2D & 2 & 0.32 & interior & 0.008 $\pm$ 0.001 & 0.011 $\pm$ 0.001 & 0.009 $\pm$ 0.001 & 0.010 $\pm$ 0.001 & 0.008 $\pm$ 0.001 & 0.023 $\pm$ 0.003 \\
Zakharov 5D & 5 & 0.33 & interior & 0.000 $\pm$ 0.000 & 0.000 $\pm$ 0.000 & 0.000 $\pm$ 0.000 & 0.000 $\pm$ 0.000 & 0.000 $\pm$ 0.000 & 0.000 $\pm$ 0.000 \\
Rastrigin 3D & 3 & 0.50 & interior & 0.844 $\pm$ 0.097 & 0.484 $\pm$ 0.059 & 0.386 $\pm$ 0.048 & 0.512 $\pm$ 0.062 & 0.519 $\pm$ 0.071 & 1.119 $\pm$ 0.059 \\
\bottomrule
\end{tabular}